\documentclass[remotesensing,article,submit,pdftex,moreauthors]{Definitions/mdpi} 
\firstpage{1} 
\pubvolume{1}
\issuenum{1}
\articlenumber{0}
\pubyear{2025}
\copyrightyear{2025}
\datereceived{ } 
\daterevised{ } 
\dateaccepted{ } 
\datepublished{ } 
\hreflink{https://doi.org/} 
\usepackage{subcaption}
\renewcommand{\linenumbers}{}

\Title{An Object-Based Deep Learning Approach for Building Height Estimation from Single SAR Images}

\TitleCitation{An Object-Based Deep Learning Approach for Building Height Estimation from Single SAR Images}

\Author{
Babak Memar\orcidA{} $^{1}$,
Luigi Russo\orcidB{} $^{1}$,
Silvia Liberata Ullo\orcidC{} $^{2}$,
and Paolo Gamba\orcidD{} $^{1}$*
}

\AuthorNames{Babak Memar, Luigi Russo, Silvia Liberata Ullo, Paolo Gamba}

\AuthorCitation{Memar, B.; Russo, L.; Ullo, S.L.; Gamba, P.}

\address{%
$^{1}$ \quad Department of Electrical, Computer and Biomedical Engineering, University of Pavia, 27100 Pavia, Italy; babak.memar01@universitadipavia.it; luigi.russo02@universitadipavia.it; paolo.gamba@unipv.it\\
$^{2}$ \quad Department of Engineering, University of Sannio, 82100 Benevento, Italy; ullo@unisannio.it
}

\corres{Correspondence: paolo.gamba@unipv.it (P. G.)}

\abstract{Accurate estimation of building heights using very high resolution (VHR) synthetic aperture radar (SAR) imagery is crucial for various urban applications. This paper introduces a Deep Learning (DL)-based methodology for automated building height estimation from single VHR COSMO-SkyMed images: an object-based regression approach based on bounding box detection followed by height estimation. This model was trained and evaluated on a unique multi-continental dataset comprising eight geographically diverse cities across Europe, North and South America, and Asia, employing a cross-validation strategy to explicitly assess out-of-distribution (OOD) generalization. The results demonstrate highly promising performance, particularly on European cities where the model achieves a Mean Absolute Error (MAE) of approximately one building story (2.20 m in Munich), significantly outperforming recent state-of-the-art methods in similar OOD scenarios. Despite the increased variability observed when generalizing to cities in other continents, particularly in Asia with its distinct urban typologies and prevalence of high-rise structures, this study underscores the significant potential of DL for robust cross-city and cross-continental transfer learning in building height estimation from single VHR SAR data.}


\begin{document}

\section{Introduction} \label{sec:intro}
The integration of remote sensing (RS) techniques with Deep Learning (DL) method\nobreak olo- gies has emerged as a highly effective approach for detecting two-dimensional (2D) and three-dimensional (3D) building feature extraction~\cite{10301648}~\cite{XIAO2023103258}. Still,  the accurate estimation of 3D changes in urban environments remains a challenging task due to factors such as obstruction from neighboring structures, variations in building materials, and complex urban geometries~\cite{rs16050740}~\cite{Hao2024}. 

More recent approaches for 3D building height estimation can be categorized into three main groups according to the type of imagery utilized:
\begin{itemize}
\item Multimodal fusion techniques that combine radar and optical data to enable the simultaneous extraction of complementary features from both image sources. This integration allows for a geometric analysis of the interactions between sunlight, buildings, and shadows in optical imagery, while also taking advantage of the all-weather, day-and-night imaging capabilities of radar data. 
\item Methodologies that relies on a time sequence of data sets coming exclusively from a single data source, either radar or optical imagery. This strategy is motivated by the desire to simplify the preprocessing workflow, reduce computational and storage overhead, and minimize the potential for errors that can arise during data fusion.
\item Methodologies that rely solely on a single image from a single data source, aiming at the best trade-off between the simplicity of data acquisition and the detailed structural information necessary for precise analysis.
\end{itemize}
With respect to the first group of approaches, a good example is \cite{CAI2023103399}, which introduces the Building Height Estimating Network (BHE-NET) to process Sentinel-1 and Sentinel-2 in parallel. The network is based on the U-Net architecture \cite{ronneberger2015u}, which serves as the core framework for the model. Similarly, in \cite{10283039} the authors proposed a methodology, the Multimodal Building Height Regression Network (MBHR-Net), a DL model specifically designed to process and analyze data from the same sources. Conversely, \cite{rs16060958} introduces a core innovation relying in the multi-level cross-fusion approach, which integrates features from SAR and electro-optical (EO) images. The model incorporates semantic information, which is particularly beneficial in densely built environments, as it helps to differentiate building structures and their respective heights.
Finally, the work in~\cite{YADAV2025114556} introduces a so called Temporally Attentive and Swin Transformer-enhanced dual-task UNet model (T-SwinUNet) to simultaneously perform feature extraction, building height estimation, and building segmentation. The model leverages complex, multi-modal time series data - specifically, 12 months of Sentinel-1 SAR and Sentinel-2 MSI imagery, including both ascending and descending passes - to capture local and global spatial patterns.
\\Among the second group of approaches, Ref.~\cite{LI2020111705} introduces a novel methodology to estimate building heights using only Sentinel-1 Ground Range Detected (GRD) data. By utilizing the VVH indicator in combination with reference building height data derived from airborne LiDAR images, the authors developed a calibrated building height estimation model, grounded on reliable reference points from seven U.S. cities. As another example, in ~\cite{buildings14113571}  the limitation of satellite-based altimetry methodologies to distinguish between the photons belonging to buildings and other objects is discussed.
\\In the third group, some methodologies leverage the grouping of 2D primitives, such as edges and segments, to infer building heights and delineate urban structures \cite{CHAMPION20101138}. However, these approaches generally rely on additional auxiliary information, such as shadows and wall evidence, to infer building heights. 
More recent works, \cite{RECLA2024104} and \cite{SUN202279}, demonstrate the feasibility of estimating building heights from single VHR SAR images. In particular, Ref.~\cite{RECLA2024104} proposes a DL model trained on over $50$ TerraSAR-X images from eight cities for reconstructing urban height maps from single VHR SAR images by integrating sensor-specific knowledge. Their method converts SAR intensity images into shadow maps to filter out low-intensity pixels, then applies mosaicking and filtering to refine height estimates. Complementarily, Ref.~\cite{SUN202279} introduces an object-based approach that estimates building heights via bounding-box regression between the footprint  and building bounding boxes, an instance-based formulation particularly effective for structures with clearly defined footprints. 

Building upon the strategies as described before, this paper aims to harness the detailed information present in single VHR SAR images to deliver accurate and efficient 3D urban structure analysis without the complexities of multi-source data fusion or multi-temporal analysis. An object-based approach is thus proposed, utilizing footprint geometries to perform instance-based height estimation. The method exploits building boundaries such as Sun et al. \cite{SUN202279}, but introduces several key distinctions.  First, while Sun et al. employ four input features, including the bounding box center coordinates, this approach uses only the bounding box dimensions (length and width), as the center coordinates contribute marginally to height prediction when the box is well aligned with the footprint. This simplification reduces model complexity and the risk of overfitting to spatial patterns.
Second, unlike~\cite{SUN202279}, the methodology presented in this paper explicitly operates in the ground range domain, avoiding the need to reproject data into the SAR image (slant range) plane. Instead of computing bounding boxes aligned with the slant-range geometry, as in Sun et al., our method accounts for the orbit geometry and incidence angle by directly rotating and adjusting the footprint-aligned bounding box (FBB) in ground coordinates. This preserves geometric consistency with the SAR acquisition geometry while simplifying the processing chain.

To test the proposed methodology, this study systematically evaluates the strengths and limitations of the proposed approach across different urban morphologies. The approach is therefore applied to COSMO-SkyMed VHR radar images acquired from a geographically diverse set of cities, including Buenos Aires, Los Angeles, Milan, Munich, New York, Rome, Shanghai, and Shenzhen. Since this city set represents a wide spectrum of urban environments, the use of COSMO-SkyMed data with consistent orbital direction and acquisition mode (as detailed in Table~\ref{acquisition_modes}) ensures that a robust and generalizable model is trained.

\noindent The rest of the article is organized as follows. In Section \ref{sec:dataset}, an overview of the study area is provided along with a detailed description of the datasets, including reference data acquisition, VHR SAR image selection, and pre-processing steps. In Section \ref{sec:methodology_babak}, the proposed object-based methodology for building height estimation is presented. In Section \ref{sec:results}, a detailed analysis of the performance of the proposed method in estimating building heights for different cities is reported, and finally, Section \ref{sec:conclusions} summarizes the main findings of the study.

\section{Study area and Data}\label{sec:dataset} 
\begin{figure*}[!ht]
    \centering
    \includegraphics[width=0.9\textwidth]{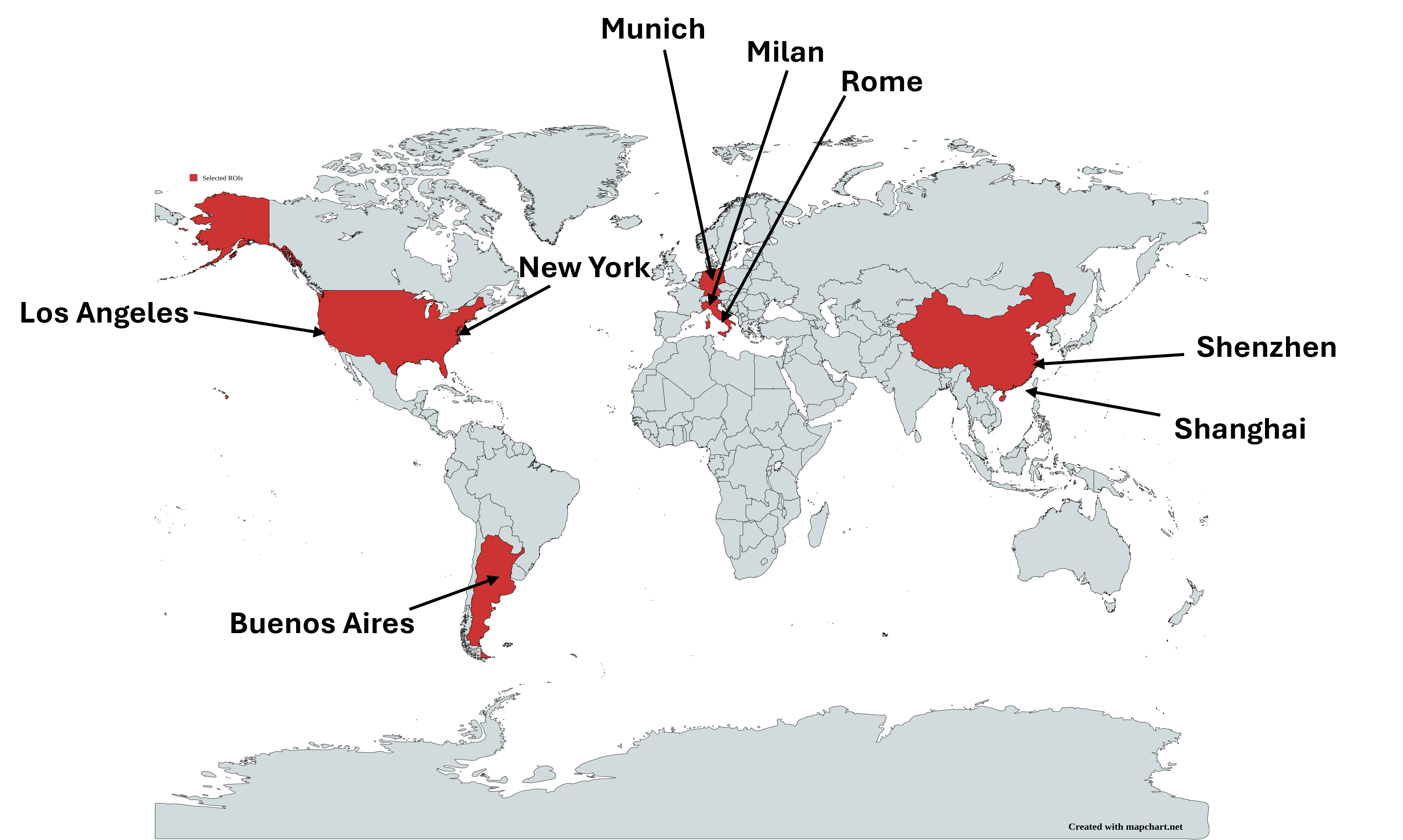}
    \caption{Study area consisting of 8 cities: Los Angeles, New York, Buenos Aires, Munich, Milan, Rome, Shanghai, and Shenzhen is highlighted in red across 3 different continents.}
    \label{study_area}
\end{figure*}

The study was conducted across eight cities located on three continents: Milan, Rome, and Munich in Europe; Shenzhen and Shanghai in Asia; and Los Angeles, New York, and Buenos Aires in the Americas. The geographical distribution of these areas is illustrated in Figure \ref{study_area}. The selection of these cities was primarily driven by the aim of developing an adaptable and scalable methodology.  Furthermore, focusing on cities across different continents ensured exposure to a wide variety of buildings with diverse characteristics. These variations stem from differences in building materials, urban development conditions, architectural structures, and even cultural distinctions across regions. 
The urban morphologies of these eight cities vary significantly, reflecting their historical development, planning strategies, and cultural influences. European cities like Milan, Rome, and Munich have predominantly low- to mid-rise structures, with historic city centers featuring irregular street patterns. Although Milan has some modern high-rise districts, Rome remains largely uniform in height due to preservation laws. Munich enforces strict height limits, maintaining a balance between tradition and modernity. In contrast, Shenzhen and Shanghai in Asia are characterized by high-rise dominance, with Shenzhen's planned grid and super-tall skyscrapers showcasing rapid urbanization, whereas Shanghai blends historic low-rise neighborhoods with its futuristic skyline in Pudong. American cities exhibit a broader spectrum of urban density and planning. New York City stands out for its extreme verticality, with some of the tallest buildings in the world densely packed in Manhattan, while Los Angeles is defined by urban sprawl, with skyscrapers concentrated in downtown but vast low-rise suburbs dominating the landscape. Buenos Aires combines both aspects, featuring a European-influenced grid with a mix of mid-rise and high-rise structures, maintaining a dense core that gradually disperses into suburban areas. As a result, differences in feature distributions can complicate model training and lead to inconsistent learning. However, an effective training data split can balance these features, thereby enhancing the model's ability to generalize across diverse urban environments.

\subsection{Input Data}\label{input_data}
The COnstellation of Small Satellites for the Mediterranean Basin Observation (COSMO-SkyMed) represents the largest Italian investment in RS space programs. It is jointly funded by the Italian Space Agency (ASI) and the Italian Ministry of Defence and has been observing the Earth since $2007$ \cite{FCOVELLO2010171}. The \href{https://earth.esa.int/eogateway/missions/cosmo-skymed}{COSMO-SkyMed (CSK)} facilitates Earth observation for both civilian and defense purposes. By capturing high-resolution images under all weather conditions, it supports a diverse range of applications, including risk management, scientific research, commercial activities, and defense operations. The four satellites of the CSK constellation follow a sun-synchronous orbit, ensuring consistent global coverage with regular revisit times. When the full constellation is operational, it can revisit a specific area on Earth within a few hours, with varying incidence angles depending on the observation requirements. CSK provides a variety of acquisition modes, each tailored to meet specific observational requirements. These modes offer flexibility in spatial resolution, coverage area, and revisit time, making the system suitable for a wide range of applications.

In this study, we used the CSK StripMap (SM) HIMAGE mode, which maintains constant radar transmit/receive configurations to capture the full Doppler bandwidth, featuring a 40 km swath width, azimuth coverage of about 40 km (for 7-second acquisitions), PRF values of 34 kHz, and chirp durations of 3540 microseconds, with chirp bandwidth ranging from 65 MHz to 140 MHz based on ground resolution. To ensure the utilization of SAR data with consistent characteristics and the highest coverage of the examined cities, the data of CSK Satellite 2 (CSKS2), acquired in descending orbital path and processed at Level-1D, were employed. This acquisition mode provides a spatial resolution of 2.5m $\times$ 2.5m, making it highly suitable for detailed urban analysis. Table \ref{acquisition_modes} briefly shows an overview of the CSK images used in this study. 
\begin{table*}[!ht]
    \centering
    \caption{Overview of COSMO-SkyMed images utilized in this study. For each city, with the exception of Los Angeles, one SAR image was selected with the same orbital pass and polarization, and the closest time interval to the corresponding reference data. In Los Angeles case, the extensive urban area and the variety of reference data led to the utilization of three SAR images. The coverage area refers to both urban and non-urban areas.}
    \resizebox{\textwidth}{!}{
        \begin{tabular}{cccccc}
            \toprule
            \textbf{City} & \textbf{Acquisition Date} & \textbf{Incidence angles (\textdegree)} & \textbf{Coverage Area (\(\text{km}^2\))} & \textbf{Polarization} & \textbf{Orbit Pass} \\
            \midrule
            Buenos Aires & 10 Jan 2021 & 26.48 & 85.56 & HH & Descending \\
            Los Angeles (North-west) & 05 Jan 2017 & 47.38 & 661.68 & HH & Descending \\ 
            Los Angeles (South-west) & 05 Jan 2017 & 47.38 & 250.28 & HH & Descending \\ 
            Los Angeles (South-east) & 04 Apr 2017 & 50.09 & 144.50 & HH & Descending \\
            Milan & 18 Nov 2022 & 30.55 & 2040.95 & HH & Descending \\
            Munich & 29 Jul 2022 & 20.04 & 2694.41 & HH & Descending \\
            New York & 05 Jan 2024 & 24.03 & 751.86 & HH & Descending \\
            Rome & 18 Sep 2023 & 24.61 & 2027.173 & HH & Descending \\
            Shanghai & 31 Aug 2018 & 30.54 & 3312.83 & HH & Descending \\
            Shenzhen & 02 Apr 2017 & 29.01 & 1865.15 & HH & Descending \\
            \bottomrule
        \end{tabular}
    \label{acquisition_modes}
    }
\end{table*}

\begin{table*}[!ht]
    \centering
    \caption{Overview of reference data for each study area. It reflects the diversity of urban environments and building heights across eight different cities in this study, includes information about the year of data collection, the number of data patches, the number of buildings categorized by height, and the data provider for each city.}
    \resizebox{0.75\textwidth}{!}{
        \begin{tabular}{ccccccc}
            \toprule
            \textbf{City} & \textbf{Year} & \textbf{\#Patches} & \multicolumn{2}{c}{\textbf{\#Buildings}} & \textbf{Provider} \\
            \cmidrule(lr){4-5}
                & & & \text{$\mathbf{h < 40}$ m} & \text{$\mathbf{h \geq 40}$ m} & \\
            \midrule
            Buenos Aires & 2021 & 579 & 544,726 & 13,888 & \href{https://data.buenosaires.gob.ar/dataset/tejido-urbano/resource/3db0ef7b-5353-4c4d-ae2c-fb1ffdba8f35}{BA Data} \\
            Los Angeles & 2017 & 3,817 & 857,521 & 578 & \href{https://geohub.lacity.org/datasets/}{Los Angeles GeoHub} \\
            Milan & 2022 & 7,929 & 252,540 & 72 & \href{https://eubucco.com/data/}{EUBUCCO} \\
            Munich & 2022 & 10,495 & 86,842 & 80 & EUBUCCO \\
            New York & 2024 & 2,643 & 719,808 & 4,212 & \href{https://opendata.cityofnewyork.us/}{NYC Open Data} \\
            Rome & 2022 & 7,802 & 303,093 & 139 & EUBUCCO \\
            Shanghai & 2017 & 12,935 & 316,935 & 24,646 & \href{https://www.research-collection.ethz.ch/handle/20.500.11850/611672}{Research Collection} \\
            Shenzhen & 2017 & 6,879 & 308,429 & 20,034 & Research Collection \\
            \bottomrule
        \end{tabular}
    \label{ref_data}
    }
\end{table*}

\subsection{Reference Data}\label{reference_data}
A significant effort in this study was dedicated to collecting reference data for model training from reliable and diverse sources, as detailed in Table \ref{ref_data}. For the European cities of Milan, Rome, and Munich, the reference data were obtained from EUBUCCO \cite{eubucco_2023}, a scientifically curated database that provides individual building footprints for over 200 million structures across the 27 European Union member states and Switzerland. The EUBUCCO database integrates information from 50 open government datasets and Open Street Map (OSM), which have been systematically collected, harmonized, and partially validated. 

The reference data for New York City were sourced from the \href{https://opendata.cityofnewyork.us/}{NYC Open Data} database, an extensive platform managed and regularly updated by the Open Data Team at the NYC Office of Technology and Innovation (OTI). This initiative focuses on documenting and managing public datasets to ensure accessibility and transparency. Similarly, the required data for Los Angeles are publicly available through the \href{https://geohub.lacity.org/datasets/}{Los Angeles GeoHub}, a centralized open data platform managed by the City of Los Angeles. For this study, data from the publicly available 2017 dataset were used. 

Reference data for urban regions in China, instead, posed significant challenges due to the limited availability of publicly accessible datasets. In this case, the data were sourced from \cite{egger2023buildingfloorspacechinadataset}. The authors employ Sentinel-1 and Sentinel-2 satellite imagery to calculate building footprints and heights for 40 major cities across China. This dataset provided a scientifically rigourous resource for urban analysis in the region of interest, although it is not validated at the same level as the othere public datasets.    

\begin{figure*}[!ht]
    \centering
    \includegraphics[width=0.9\textwidth]{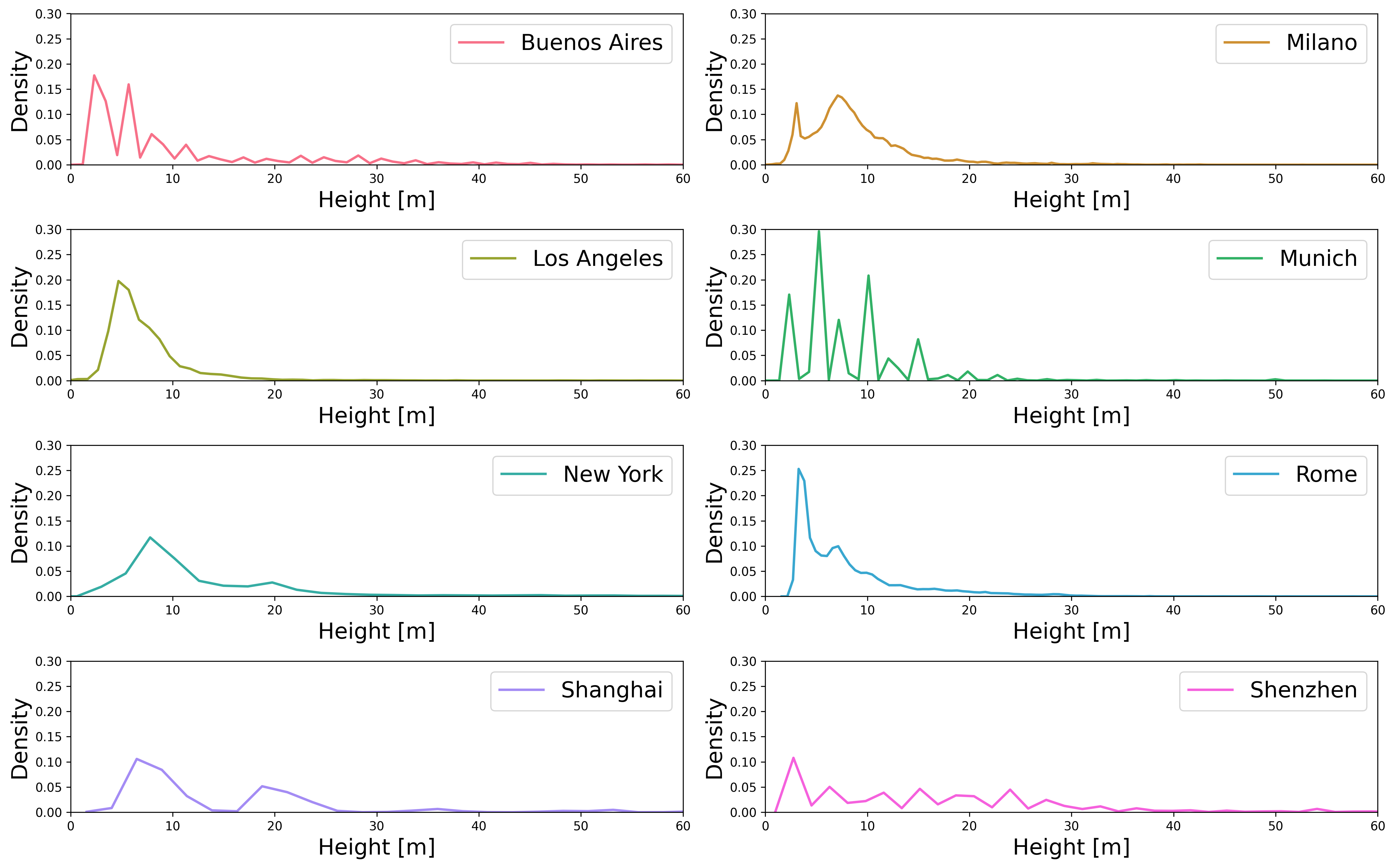}
    \caption{Building height density distributions for the eight cities considered in this study. As shown in the figure, the distributions exhibit a typical long-tail pattern, characteristic of urban environments, where lower buildings are predominant, while taller structures are less frequent.}
    \label{city_height_distributions}
\end{figure*}

\subsection{Data pre-processing and splitting}\label{subsec:preprocess}

The input data was segmented into smaller tiles using a standardized approach, to ensure consistency in data processing while allowing a rigorous evaluation of the performance of the method. More specificaly the SAR images and the corresponding reference data were partitioned into $256 \times 256$ pixel patches with a 20$\%$ overlap, ensuring spatial continuity and preventing the loss of information at the edges. 
This process yielded a total of 53,079 paired SAR and reference patches across all urban environments. Table \ref{ref_data} introduced above details the distribution of these patches city by city. 
Although the object-based method presented in the next section uses only a subset of these patches, the full dataset remains a significant outcome of this work and can support future benchmarking and evaluation of new methodologies.

\subsection{Training and Optimization}\label{subsec:training_optimization_pixel_object}
\noindent The object-based approach utilizes an individual patch for each building and its correspond\nobreak ing footprint mask as input data. In other words, each building is individually identified, and a specific image patch and corresponding footprint mask is generated. In the eight cities used in this study, a total of $3,317,956$ buildings were identified and a corresponding number of patches was generated (see Table~\ref{ref_data}). However, training the object-based model on the entire set of patches poses significant computational challenges. To address this constraint, a random subset of $20,000$ samples was selected from each city, resulting in $160,000$ training patches in total. Subsequently, the model was trained and evaluated using a cross-validation strategy, where in each run one city was held out as an out of distribution (OOD) test set while data from the remaining cities were used for training, as better detailed in Section~\ref{sec:results}. 
Since the object-based approach regresses a single height value per building and any large error in this prediction affects the overall estimate of a single building, a quadratic penalty is preferable. Therefore, the Mean Squared Error (MSE) loss was used, defined as \( \text{MSE} = \frac{1}{n} \sum_{i=1}^{n} (y_i - \hat{y}_i)^2 \), where \( y_i \) represents the ground truth values, \( \hat{y}_i \) denotes the predicted values, and \( n \) is the total number of building objects in a batch. MSE strongly penalizes large discrepancies, thus encouraging the model to correct substantial deviations more aggressively.

\section{Object-Based Building Height Estimation}\label{sec:methodology_babak}

\begin{figure*}[htp]
	\centering
	\includegraphics[width=0.6\textwidth]{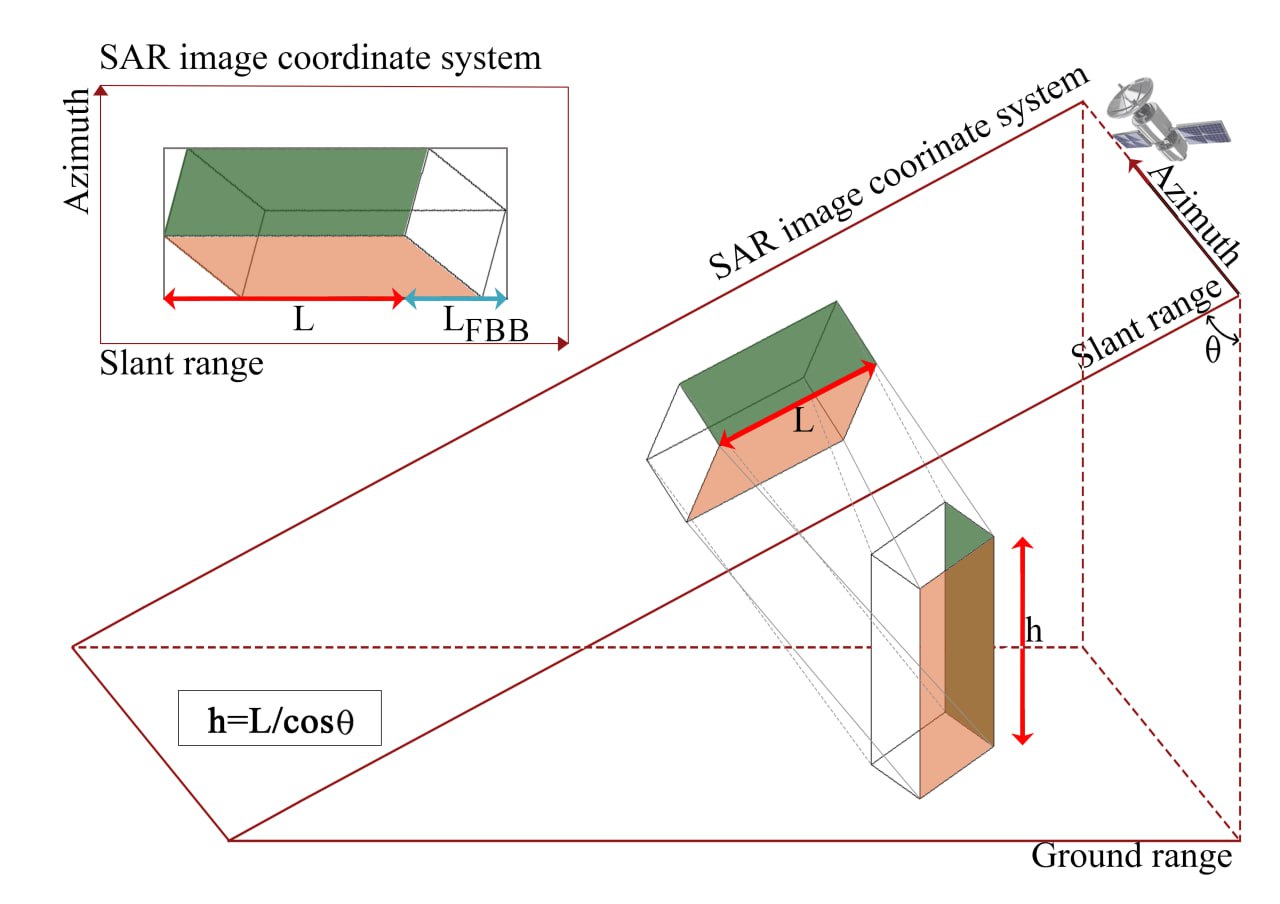}
	\caption{Graphical representation of a building in the \textit{UTM} and \textit{SAR image} coordinate systems.}
	\label{UTM_SAR_coordinateSystem}
\end{figure*}

The basic element of the object-based building analysis approach is the building footprint, which serves as a crucial source of information in various domains, including disaster management and urban planning~\cite{wangiyana2022data}. 
Therefore, a binary footprint mask is generated for each building, along with the corresponding segment of the SAR image. To achieve this, in each initial patch created during the pre-processing stage, as described in Section \ref{subsec:preprocess}, all building footprints are identified and the corresponding SAR patch is extracted. Given that the initial patches include an overlap of $20\%$, some buildings may appear two to four times in different patches. To ensure a clean and non-redundant dataset, duplicate instances are identified and removed based on the geometric properties of each building. This refinement process ensures that each building is represented only once in the final training dataset.

As mentioned, we rely on the ideas proposed in ~\cite{SUN202279}, where the authors describe the relationship between the geometry of the SAR imagery and the actual physical structure of a building on the ground. Figure \ref{UTM_SAR_coordinateSystem} illustrates this relationship. SAR imagery is acquired in a slant range coordinate system, where the projection of a building's footprint and height is inherently influenced by this geometry. 

In this way, two bounding areas can be defined based on the structural characteristics of each building. The first one is the footprint bounding box (FBB), which encompasses the building's ground footprint. The second one is the building bounding box (BBB), which includes the projection of both the footprint and the building's height within the SAR image. Therefore, these two bounding box can be used to calculate the height of the building by applying the incidence angle (\(\theta)\) of the sensor.

By delineating these two regions, the difference in length between them can be calculated as follows:

\begin{equation} \label{e:Layover}
    L = L_{\text{BBB}} - L_{\text{FBB}}
\end{equation}

\noindent where $[L_{BBB}]$ and $[L_{FBB}]$ are the lengths of the BBB and FBB, respectively, in the range direction. Thus, the building height can be estimated as:

\begin{equation} \label{e:height}
    h = L/\cos\theta
\end{equation}

\noindent where \textit{h} is the height of the building, \textit{L} is the difference in the lengths of the BBB and its corresponding FBB, and \textit{$\theta$} the incidence angle of the satellite. 

\subsection{Features Generation}\label{subsec:feature_gen}
It is important to note that it is crucial to factor in the satellite's position when monitoring a building, ensuring that the FBB and BBB are accurately calculated to achieve reliable results. By using the inclination angle, defined as the angle between the \textit{orbital plane} and the \textit{equator}, FBB and BBB can be calculated in a manner that accurately cover the entire area of the building, aligning with the satellite's angle of view.  

\begin{figure*}[htp!]
	\centering
	\includegraphics[width=1.\textwidth]{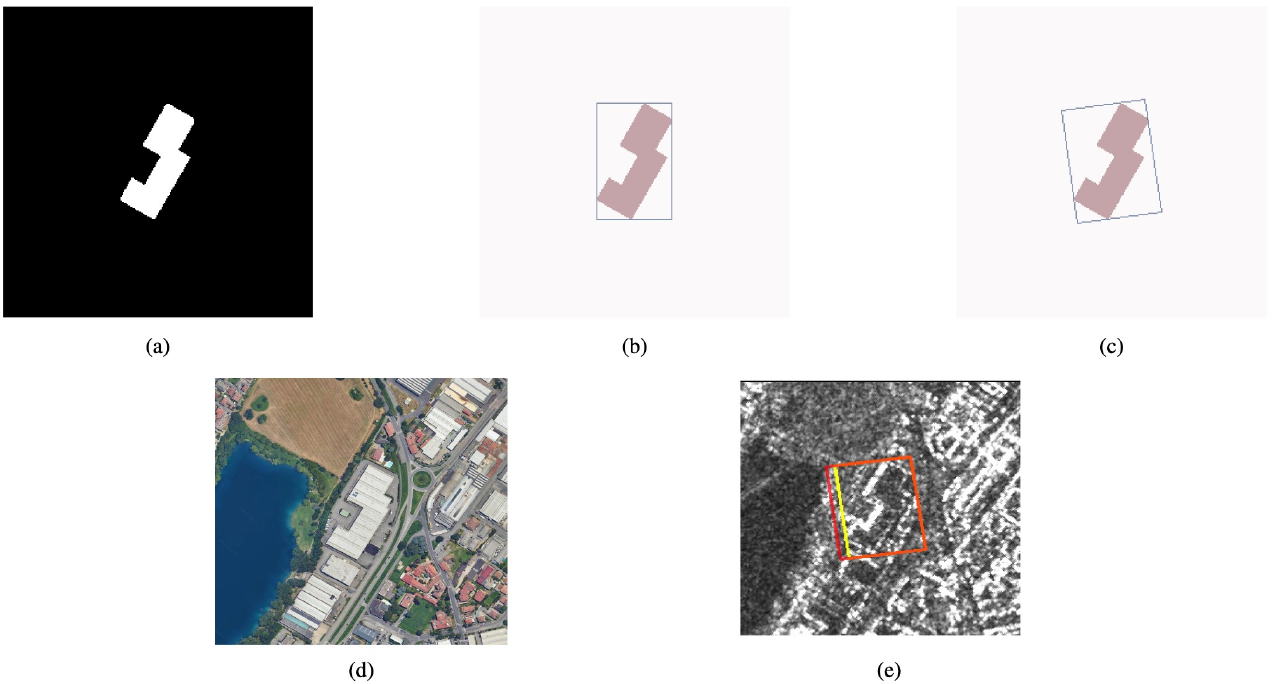}
	\caption{Extracting FBB with consideration of the inclination angle: 
    (a) building footprint in New York City, (b) FBB before and (c) after angle correction, and the corresponding (d) optical and (e) SAR images with superimposed FBB and BBB (yellow and red boxes, overlap in orange).}
	\label{fig:fbb_bbb}
\end{figure*}

The FBB is extracted by looking for the smallest enclosing rectangle that fully covers the footprint area. Then, this rectangle is rotated according to the inclination angle, and its size is adjusted to ensure complete coverage of the footprint. Figure ~\ref{fig:fbb_bbb} represents the process of extracting the FBB for a sample building footprint.

\subsection{Model Architecture}\label{subsec:network_architecture_babak}
\begin{figure*}[htp!]
	\centering
	\includegraphics[width=1.\textwidth]{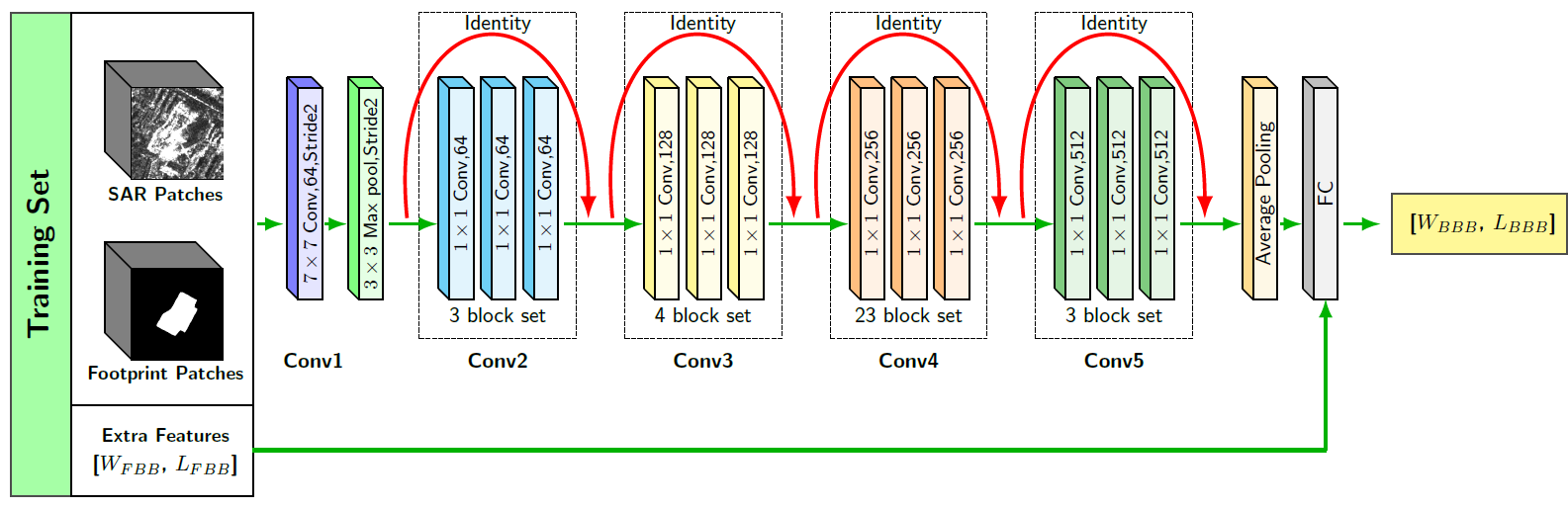}
	\caption{Overview of the proposed workflow for the object-based building height estimation. The model utilizes ResNet101 as the backbone in combination with the extracted features from 2D coordinates of the building. 
    After generating the extra features using the footprint of the buildings, the model concatenates a SAR image and the corresponding footprint from the training dataset. Using the extracted features and the 2D coordinates information, the network predicts the building projected footprint to estimate the height of the building.}
	\label{fig:ResNet101}
\end{figure*}

The overall framework of the instance-based methodology is depicted in Figure \ref{fig:ResNet101}, illustrating the sequential processes involved in detecting and analyzing individual objects within the dataset. The proposed methodology consists of three main parts:
\begin{enumerate}
    \item a DL framework that extract the features from the input SAR image and its corresponding footprint mask;
    \item a feature generator that calculates the extra attributes for each building separately;
    \item a fully connected layer to integrate the extracted features and extra attributes to estimate the height of the buildings.
\end{enumerate}

The DL framework is based on ResNet-101 ~\cite{he2016deep}. ResNet-101 is a deep residual network comprising 101 weighted layers, structured into five convolutional blocks, each containing multiple residual sub-networks. This architecture is specifically designed to facilitate efficient feature extraction by employing residual learning, which addresses the vanishing gradient and degradation problems commonly encountered in very deep networks.

In the proposed methodology, all the layers of ResNet-101 are utilized to extract meaningful feature representations from SAR imagery. The model initially takes the SAR image and building footprint as input. In the first stage, convolutional layers process these inputs to generate an initial set of feature maps. Then, additional features, which derived from FBB using the feature generator component, is integrated with the extracted feature maps before being passed to the fully connected layer. The output of the fully connected layer is then compared with the ground truth data in a regression framework, where the objective is to estimate the bounding boxes that define building extents. Finally, from these bounding boxes, the building height is computed.

\begin{figure*}[htp]
	\centering
	\includegraphics[width=0.7\textwidth]{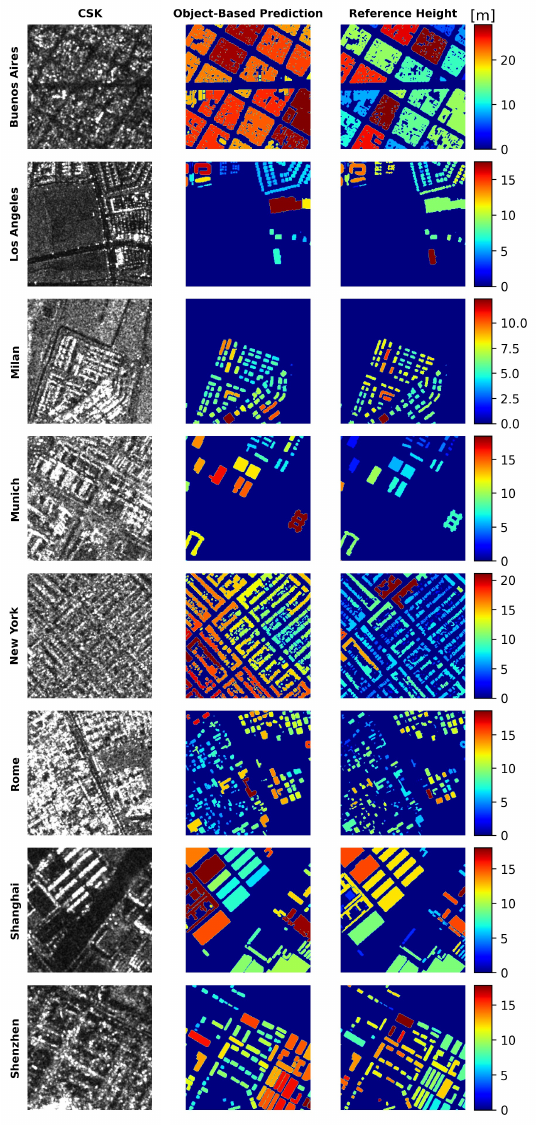}
        \caption{Comparison of building height estimation results for different cities. Each row displays an image patch for a specific city (from top to bottom: Buenos Aires, Los Angeles, Milan, Munich, New York, Rome, Shanghai, and Shenzhen), showing (from left to right): the CSK input SAR patch,the object-based height estimation, and the corresponding ground truth height map. The predictions are OOD, computed from models trained on all the other cities.}
	\label{fig:patch_comparisons}
\end{figure*}

\section{Results and Discussion}\label{sec:results}
\begin{table*}[!ht]
    \centering
    \caption{Quantitative evaluation of the object-based building height estimates across the eight different unseen cities for each run.}
    \small
    \renewcommand{\arraystretch}{1.2} 
    \begin{tabular}{ccccccc}
        \toprule
        \textbf{City} & \multicolumn{3}{c}{\textbf{MAE [m]}} & \multicolumn{3}{c}{\textbf{RMSE [m]}} \\
        \cmidrule(lr){2-4} \cmidrule(lr){5-7}
                         & \textbf{$\forall$ h} & \textbf{$\mathbf{h < 40}$ m} & \textbf{$\mathbf{h \geq 40}$ m} & \textbf{$\forall$ h}  & \textbf{$\mathbf{h < 40}$ m} & \textbf{$\mathbf{h \geq 40}$ m} \\
        \midrule
        {Buenos Aires} & 8.66 & 7.79 & 43.15 & 13.43 & 11.44 & 46.15 \\ \hline
        {Los Angeles} & 2.24 & 2.21 & 53.36 & 3.29 & 3.02 & 59.71 \\ \hline
        {Milan} & 2.26 & 2.25 & 37.91 & 3.67 & 3.61 & 39.01 \\ \hline
        {Munich}  & 2.20 & 2.16 & 61.77 & 3.97 & 3.46 & 74.42 \\ \hline
       {New York} & 7.43 & 7.26 & 36.19 & 9.36 & 8.79 & 42.93 \\ \hline
        {Rome} & 10.71 & 10.70 & 20.02 & 12.81 & 12.80 & 30.52 \\ \hline
        {Shanghai} & 10.00 & 8.08 & 34.68 & \textbf{14.89} & 10.51 & 40.64 \\ \hline
        {Shenzhen}  & 10.51 & 8.31 & 44.32 & 16.35 & 10.58 & 51.57 \\ \hline
        \bottomrule
    \end{tabular}
    \label{compare_results}
\end{table*}

\begin{table*}[!ht]
    \centering
    \caption{Quantitative evaluation of the object-based building height estimation methods for the in-distribution (ID) test, where the entire dataset (without any geographic split) was divided into $70\%$ for training and $30\%$ for validation.}
    \small
    \renewcommand{\arraystretch}{1.2} 
    \resizebox{\textwidth}{!}{%
        \begin{tabular}{ccccccc}
            \toprule
            \textbf{Test} & \multicolumn{3}{c}{\textbf{MAE [m]}} & \multicolumn{3}{c}{\textbf{RMSE [m]}} \\
            \cmidrule(lr){2-4} \cmidrule(lr){5-7}
            & \textbf{$\forall$ h} & \textbf{$\mathbf{h < 40}$ m} & \textbf{$\mathbf{h \geq 40}$ m} & \textbf{$\forall$ h}  & \textbf{$\mathbf{h < 40}$ m} & \textbf{$\mathbf{h \geq 40}$ m} \\
            \midrule
            In-Distribution (70-30) & 4.95 & 4.28 & 35.63 & 8.87 & 6.57 & 41.09 \\
            \bottomrule
        \end{tabular}
    }
    \label{compare_7030}
\end{table*}

In order to evaluate and compare the performance of the proposed methodology, the Mean Absolute Error (MAE) and Root Mean Squared Error (RMSE) were utilized as evaluation metrics, and a series of experiments was performed. 

\begin{figure*}[htp]
	\centering
	\includegraphics[width=1\textwidth]{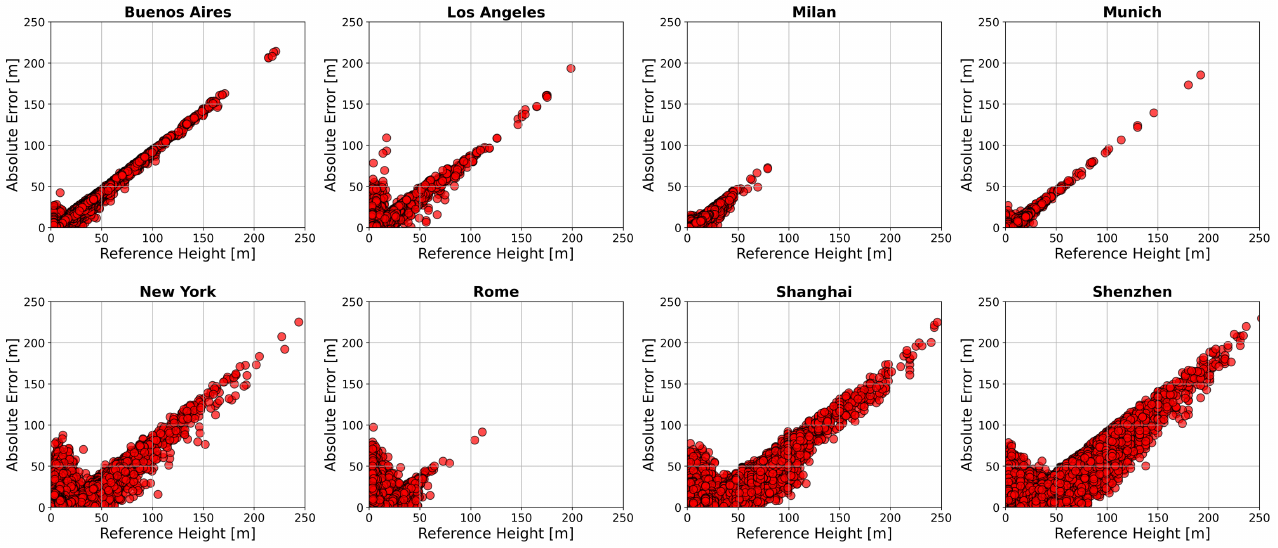}
	\caption{Scatter plots of absolute errors per city relative to reference heights. Each point represents the error of a single building.}
	\label{fig:scattergram_eight_cities}
\end{figure*}

In the first set of experiments a cross-validation strategy was considered. More specifically, the data from seven cities served as the training dataset, while the remaining city was used for testing and related evaluation, resulting in eight different experiments. This strategy ensures a comprehensive assessment of the generalization capabilities of the proposed methodology on unseen urban environments. The quantitative results are presented in Table \ref{compare_results}. MAE and RMSE were computed across three scenarios: for all buildings, for buildings with a reference height below 40 meters, and for buildings above 40 meters. This classification enables a deeper analysis of the performance over different building height ranges, offering a more comprehensive assessment of the obtained precision in both low-rise and high-rise urban contexts. Qualitative examples illustrating these results for several cities are also shown Figure~\ref{fig:patch_comparisons}.
\\The results reveal a complex interplay between methodology and urban environment. In Buenos Aires, the relatively poor performance of the method is likely attributed to the exceptionally high building density, a factor that complicates object identification and delineation in SAR imagery. The extensive overlap of structures poses a significant challenge for any object-based methodology, which relies on accurate building boundary box detection. A similar behavior is observed for New York. Here a contributing factor is the exceptionally high density of very tall buildings within specific areas of the city. For instance, Manhattan exhibits one of the highest densities of buildings exceeding 40 meters in height, a feature that is not equally present in the other cities analyzed in this study. 
\\Conversely, the proposed approach demonstrates good performances in Los Angeles. Los Angeles's urban landscape, characterized by more distinct building separation compared to Buenos Aires, likely facilitates more accurate object delineation, thus benefiting the object-based approach.
In Milan, the approach exhibits excellent performances (MAE: 2.26 m, RMSE: 3.67 m), suggesting that in cities with moderate building density and complexity, ibjecxt based approaches are working well.
However, it struggles in cities with a high prevalence of tall buildings and complex architectural diversity, such as Shenzhen (MAE: 10.51 m, RMSE: 16.35 m) and Shanghai. The uneven distribution of building heights in the training data, skewed towards lower buildings (as illustrated in Figure \ref{city_height_distributions}), likely hindered the models' ability to generalize effectively to these cities with a high proportion of skyscrapers. Furthermore, the intricate architectural styles and high building density in these Asian megacities introduce significant challenges for accurate height estimation.
\\In the second set of experiments, the dataset comprising 8 cities was divided into training and test sets with a $70\%$ to $30\%$ ratio. The quantitative results for this $70-30$ in-distribution test, illustrated in Table \ref{compare_7030}, show that the proposed methodology achieves a global MAE between 4 and 5 m, underscoring its ability to provide reasonably accurate building height estimates on a global scale. 
\\Figure \ref{fig:scattergram_eight_cities} presents a detailed spatial analysis of the absolute errors of the height estimates across eight cities, providing insights that extend beyond the aggregate metrics reported in Table \ref{compare_results}.
\\The high presence of orange regions with relatively small errors, particularly in areas corresponding spatially to buildings under 40 meters (as confirmed by visual inspection and the building height distribution in Table \ref{ref_data}), strongly suggests a consistent and reliable performance for the most common building stock. This spatial agreement in relatively low error magnitudes reinforces the quantitative findings of robust performance in lower structures.
\\However, the figure also highlights notable areas of divergence. The increased presence of yellow and red zones in regions associated with taller and potentially more complex building structures visually confirms the challenges that the method encounter with increasing height, aligning with the higher error values reported in Table \ref{compare_results} for taller buildings.

\begin{table}[!ht]
    \centering
    \caption{Comparison of the proposed model with state-of-the-art methodology applied to some of the cities belonging to our dataset.}
    \small
    \renewcommand{\arraystretch}{1.2} 
        \begin{tabular}{cccc}
            \toprule
            \textbf{City} & \textbf{Methodology} & {\textbf{MAE [m]}} & {\textbf{RMSE [m]}} \\
            \midrule
            \multirow{2}{*}{\textbf{Los Angeles}} & Kaya \cite{buildings14113571} & 4.66 & 6.42 \\ \cline{2-4}
                            & \(\text{this paper}\) & 2.24 & 3.29 \\
            \midrule
            
            \multirow{2}{*}{\textbf{Milan}} & T-SwinUNet \cite{YADAV2025114556} & 3.54 & 4.54 \\ \cline{2-4}
                            & \(\text{this paper}\) & 2.26 & \text{3.67} \\

            \midrule
            
            \multirow{2}{*}{\textbf{New York}} & Kaya  \cite{buildings14113571} & 6.24 & 8.28 \\ \cline{2-4}
                & \(\text{this paper}\) & 7.43 & 9.36 \\
            \midrule
            
            \multirow{2}{*}{\textbf{Shenzhen}} & BHEPD \cite{CHEN2023113802} & \textbf{4.00} & \textbf{6.72}  \\ \cline{2-4}
                            & \(\text{this paper}\) & 10.51 & 16.35 \\
            \bottomrule
        \end{tabular}
    \label{compare_others}
\end{table}

\noindent As a final test, the results of the proposed methodology are compared to state-of-the-art approaches i the same cities in Table \ref{compare_others}. It is worth noting that, given the range and diversity of cities considered in this study, no single reference could serve as a universal benchmark across all experimental settings. Accordingly, compatisons were performed on a city basis, by looking at the best reported results for specific citied by means of other scientifically validated techniques and input data sets.

In Los Angeles, the proposed model can be compared with, and actually outperforms, the results in \cite{buildings14113571}. Indeed, it achieved an MAE of 2.24 m vs a reported 4.66 m in the above-mentioned study, as well as an RMSE of 3.29 m vs 6.42 m. In contrast, for New York City, the performance was slightly worse (MAE: 7.43 m vs 6.24 m and RMSE: 9.36 m vs 8.28 m).
\\In the case of Milan, the T-SwinUNet model from \cite{YADAV2025114556} was used for comparison. Yadav et al. report a MAE of 3.54 m and an RMSE of 4.54 m, whereas the proposed method yields better results (MAE: 2.26 m; RMSE: 3.67 m). 
It must be noted that \cite{YADAV2025114556} exploits multimodal and multitemporal data from Sentinel-1 and Sentinel-2, hence incorporating additional input information and spatio-temporal attention mechanisms. Nevertheless, our method brings an improvement that can be partly attributed to the use of VHR SAR images (COSMO-SkyMed) vs. Sentinel-1. 
\\The last city for which we compared the extraction results with those reported in scientific literature is Shenzhen. We used for comparison the study in \cite{CHEN2023113802}, where the authors developed an effective Building Height Estimation method by synergizing Photogrammetry and Deep learning methods (BHEPD) to leverage Chinese multi-view GaoFen-7 (GF-7) images for high-resolution building height estimation. The results indicate that BHEPD outperforms our model, achieving better MAE amd RMSE values, 4.00 m and 6.72 m, respectively. This represents the worst performance of the proposed method used in this research and was left for a fair comparison and evaluation. This result (in disagreement with the other ones) can be attributed to the unique characteristics of Shenzhen: as shown in Figure \ref{city_height_distributions}, this city has a significantly high frequency of buildings taller than 40 meters. Consequently, when this city is excluded from the training set and used as test data (recall that our experiments are OOD), the model has a limited exposure to tall building heights, resulting in poorer height estimates. Furthermore, it is essential to consider the impact of building footprint dimensions on the accuracy of height estimation in the object-based methodology. In these urban settings, it is common to encounter very tall buildings, such as skyscrapers, with narrow footprints, which can complicate the estimation process in SAR (i.e., slanted) images. 
\\This city-specific comparison well describes how the proposed methodology performs under various urban conditions across different cities. In Los Angeles and New York, it demonstrates competitive accuracy relative to benchmark studies. In Milan, the use of VHR SAR images yields improved accuracy compared to state-of-the-art models that rely on multimodal and multitemporal composites, while in Shenzhen the prevalence of tall buildings with narrow footprints poses big challenges and limits to the model performance.

\section{Conclusions}\label{sec:conclusions}
This study presents and thoroughly evaluates an object-based methodology for the challenging task of building height estimation from single VHR SAR imagery. The main goals of this paper were to reduce time and computational costs using single time acquisitions VHR SAR images for each city and to train a scalable model to be applied across different continents with a reasonable amount of error. 
\\ Eight test sites (Buenos Aires, Los Angeles, Milan, Munich, New York, Rome, Shanghai, and Shenzhen) were selected in order to provide a diverse training and test sets, due to different built-up structure distributions and different cultural and historical urban settings. 
\\The experiments provided transparent critical insights into the strengths and weaknesses of the proposed approach under realistic conditions, especially in the OOD experiments.

A key highlight of this research is the strong generalization capability demonstrated for the European cities. Indeed, achieving an accuracy of approximately one building story in an out-of-distribution setting, surpassing that of recent methodologies, underscores the effectiveness of the proposed DL frameworks. This performance demonstrates its potential for deployment in urban environments that are geographically and architecturally similar, even when city-specific training data are not available.
While performance on North American and South American cities remained competitive, the noticeable decrease in accuracy observed for much more recently built cities in China (Shanghai and Shenzhen) points to the challenges in achieving robust cross-continental generalization. Factors such as distinct architectural styles, varying building densities, and the prevalence of very high-rise structures appear to pose a greater challenge for current methodologies. This suggests that while the foundational framework shows significant promise for transfer learning, further research is needed to bridge the gap in performance across highly diverse global urban landscapes.

Looking ahead, future work could focus on enhancing the model ability to generalize across diverse global urban landscapes, particularly in cities with a high prevalence of tall buildings and complex architectural diversity. Furthermore, it would be interesting to compare first and then possibly combine object-based and pixel-based methodologies to exploit their respective strengths. For generalization in specific urban landscapes, the use of fine-tuning techniques could also be explored to adapt pre-trained models to new cities more effectively.

\authorcontributions{Conceptualization, P.G., S.U., B.M., and L.R.; methodology,  B.M., and L.R.;  software, B.M., and L.R.;  validation, B.M., and L.R.;  formal analysis, P.G. and S.U.; data curation, B.M., and L.R.;  writing---original draft preparation, B.M., and L.R.;  writing---review and editing, P.G. and S.U.; visualization, L.R.; supervision, P.G.; funding acquisition, P.G. All authors have read and agreed to the published version of the manuscript.}

\funding{This research was performed in the framework of a PhD funded by NextGenerationEU, Action 4, DM n. 118, 02/03/2023. 
}

\dataavailability{The data is available to the authors by a research agreement with ASI, and can be provided upon request to other researchers after a similar agreement between their institution(s) and ASI is signed.} 

\conflictsofinterest{The authors declare no conflicts of interest.}

\reftitle{References}
\bibliography{references}

\begin{thebibliography}{999}

\bibitem[Gomroki et~al.(2023)Gomroki, Hasanlou, and Chanussot]{10301648}
Gomroki, M.; Hasanlou, M.; Chanussot, J.
\newblock {Automatic 3D Multiple Building Change Detection Model Based on Encoder–Decoder Network Using Highly Unbalanced Remote Sensing Datasets}.
\newblock {\em IEEE Journal of Selected Topics in Applied Earth Observations and Remote Sensing} {\bf 2023}, {\em 16},~10311--10325.
\newblock {\url{https://doi.org/10.1109/JSTARS.2023.3328561}}.

\bibitem[Xiao et~al.(2023)Xiao, Cao, Tang, Zhang, and Chen]{XIAO2023103258}
Xiao, W.; Cao, H.; Tang, M.; Zhang, Z.; Chen, N.
\newblock {3D urban object change detection from aerial and terrestrial point clouds: A review}.
\newblock {\em International Journal of Applied Earth Observation and Geoinformation} {\bf 2023}, {\em 118},~103258.
\newblock {\url{https://doi.org/https://doi.org/10.1016/j.jag.2023.103258}}.

\bibitem[He et~al.(2024)He, Cheng, Wang, Ren, Zhang, and Zhang]{rs16050740}
He, J.; Cheng, Y.; Wang, W.; Ren, Z.; Zhang, C.; Zhang, W.
\newblock {A Lightweight Building Extraction Approach for Contour Recovery in Complex Urban Environments}.
\newblock {\em Remote Sensing} {\bf 2024}, {\em 16}.

\bibitem[Hao et~al.(2024)Hao, Chen, Lin, Zhang, and Zheng]{Hao2024}
Hao, M.; Chen, S.; Lin, H.; Zhang, H.; Zheng, N.
\newblock {A prior knowledge guided deep learning method for building extraction from high-resolution remote sensing images}.
\newblock {\em Urban Informatics} {\bf 2024}, {\em 3},~6.
\newblock {\url{https://doi.org/10.1007/s44212-024-00038-8}}.

\bibitem[Cai et~al.(2023)Cai, Shao, Huang, Zhou, and Fang]{CAI2023103399}
Cai, B.; Shao, Z.; Huang, X.; Zhou, X.; Fang, S.
\newblock {Deep learning-based building height mapping using Sentinel-1 and Sentinel-2 data}.
\newblock {\em International Journal of Applied Earth Observation and Geoinformation} {\bf 2023}, {\em 122},~103399.
\newblock {\url{https://doi.org/https://doi.org/10.1016/j.jag.2023.103399}}.

\bibitem[Ronneberger et~al.(2015)Ronneberger, Fischer, and Brox]{ronneberger2015u}
Ronneberger, O.; Fischer, P.; Brox, T.
\newblock {U-Net: Convolutional networks for biomedical image segmentation}.
\newblock In Proceedings of the Medical Image Computing and Computer-Assisted Intervention--MICCAI 2015: 18th International Conference, Munich, Germany, October 5-9, 2015, Proceedings, Part III 18. Springer,  2015, pp. 234--241.

\bibitem[Nascetti et~al.(2023)Nascetti, Yadav, and Ban]{10283039}
Nascetti, A.; Yadav, R.; Ban, Y.
\newblock {A CNN Regression Model to Estimate Buildings Height Maps Using Sentinel-1 SAR and Sentinel-2 MSI Time Series}.
\newblock In Proceedings of the IGARSS 2023 - 2023 IEEE International Geoscience and Remote Sensing Symposium,  2023, pp. 2831--2834.
\newblock {\url{https://doi.org/10.1109/IGARSS52108.2023.10283039}}.

\bibitem[Ma et~al.(2024)Ma, Zhang, Guo, Zhou, and Geng]{rs16060958}
Ma, C.; Zhang, Y.; Guo, J.; Zhou, G.; Geng, X.
\newblock {FusionHeightNet: A Multi-Level Cross-Fusion Method from Multi-Source Remote Sensing Images for Urban Building Height Estimation}.
\newblock {\em Remote Sensing} {\bf 2024}, {\em 16}.
\newblock {\url{https://doi.org/10.3390/rs16060958}}.

\bibitem[Yadav et~al.(2025)Yadav, Nascetti, and Ban]{YADAV2025114556}
Yadav, R.; Nascetti, A.; Ban, Y.
\newblock {How high are we? Large-scale building height estimation at 10 m using Sentinel-1 SAR and Sentinel-2 MSI time series}.
\newblock {\em Remote Sensing of Environment} {\bf 2025}, {\em 318},~114556.
\newblock {\url{https://doi.org/https://doi.org/10.1016/j.rse.2024.114556}}.

\bibitem[Li et~al.(2020)Li, Zhou, Gong, Seto, and Clinton]{LI2020111705}
Li, X.; Zhou, Y.; Gong, P.; Seto, K.C.; Clinton, N.
\newblock {Developing a method to estimate building height from Sentinel-1 data}.
\newblock {\em Remote Sensing of Environment} {\bf 2020}, {\em 240},~111705.
\newblock {\url{https://doi.org/https://doi.org/10.1016/j.rse.2020.111705}}.

\bibitem[Kaya(2024)]{buildings14113571}
Kaya, Y.
\newblock {Automated Estimation of Building Heights with ICESat-2 and GEDI LiDAR Altimeter and Building Footprints: The Case of New York City and Los Angeles}.
\newblock {\em Buildings} {\bf 2024}, {\em 14}.
\newblock {\url{https://doi.org/10.3390/buildings14113571}}.

\bibitem[Champion et~al.(2010)Champion, Boldo, Pierrot-Deseilligny, and Stamon]{CHAMPION20101138}
Champion, N.; Boldo, D.; Pierrot-Deseilligny, M.; Stamon, G.
\newblock {2D building change detection from high resolution satelliteimagery: A two-step hierarchical method based on 3D invariant primitives}.
\newblock {\em Pattern Recognition Letters} {\bf 2010}, {\em 31},~1138--1147.
\newblock Pattern Recognition in Remote Sensing, {\url{https://doi.org/https://doi.org/10.1016/j.patrec.2009.10.012}}.

\bibitem[Recla and Schmitt(2024)]{RECLA2024104}
Recla, M.; Schmitt, M.
\newblock {The SAR2Height framework for urban height map reconstruction from single SAR intensity images}.
\newblock {\em ISPRS Journal of Photogrammetry and Remote Sensing} {\bf 2024}, {\em 211},~104--120.
\newblock {\url{https://doi.org/https://doi.org/10.1016/j.isprsjprs.2024.03.023}}.

\bibitem[Sun et~al.(2022)Sun, Mou, Wang, Montazeri, and Zhu]{SUN202279}
Sun, Y.; Mou, L.; Wang, Y.; Montazeri, S.; Zhu, X.X.
\newblock {Large-scale building height retrieval from single SAR imagery based on bounding box regression networks}.
\newblock {\em ISPRS Journal of Photogrammetry and Remote Sensing} {\bf 2022}, {\em 184},~79--95.
\newblock {\url{https://doi.org/https://doi.org/10.1016/j.isprsjprs.2021.11.024}}.

\bibitem[F.Covello et~al.(2010)F.Covello, Battazza, Coletta, Lopinto, Fiorentino, Pietranera, Valentini, and Zoffoli]{FCOVELLO2010171}
F.Covello.; Battazza, F.; Coletta, A.; Lopinto, E.; Fiorentino, C.; Pietranera, L.; Valentini, G.; Zoffoli, S.
\newblock {COSMO-SkyMed an existing opportunity for observing the Earth}.
\newblock {\em Journal of Geodynamics} {\bf 2010}, {\em 49},~171--180.
\newblock WEGENER 2008 - Proceedings of the 14th General Assembly of Wegener, {\url{https://doi.org/https://doi.org/10.1016/j.jog.2010.01.001}}.

\bibitem[{Milojevic-Dupont, Nikola and Wagner, Felix} et~al.(2023){Milojevic-Dupont, Nikola and Wagner, Felix}, Nachtigall, Hu, Br{\"u}ser, Zumwald, Biljecki, Heeren, Kaack, Pichler, and Creutzig]{eubucco_2023}
{Milojevic-Dupont, Nikola and Wagner, Felix}.; Nachtigall, F.; Hu, J.; Br{\"u}ser, G.B.; Zumwald, M.; Biljecki, F.; Heeren, N.; Kaack, L.H.; Pichler, P.P.; Creutzig, F.
\newblock {EUBUCCO v0.1: European building stock characteristics in a common and open database for 200+ million individual buildings}.
\newblock {\em Scientific Data} {\bf 2023}, {\em 10},~147.
\newblock {\url{https://doi.org/10.1038/s41597-023-02040-2}}.

\bibitem[Egger et~al.(2023)Egger, Rao, and Papini]{egger2023buildingfloorspacechinadataset}
Egger, P.; Rao, S.X.; Papini, S.
\newblock {Building Floorspace in China: A Dataset and Learning Pipeline},  2023,  \href{http://arxiv.org/abs/2303.02230}{{\normalfont [arXiv:cs.CV/2303.02230]}}.

\bibitem[Wangiyana et~al.(2022)Wangiyana, Samczy{\'n}ski, and Gromek]{wangiyana2022data}
Wangiyana, S.; Samczy{\'n}ski, P.; Gromek, A.
\newblock {Data augmentation for building footprint segmentation in SAR images: an empirical study}.
\newblock {\em Remote Sensing} {\bf 2022}, {\em 14},~2012.

\bibitem[He et~al.(2016)He, Zhang, Ren, and Sun]{he2016deep}
He, K.; Zhang, X.; Ren, S.; Sun, J.
\newblock Deep residual learning for image recognition.
\newblock In Proceedings of the Proceedings of the IEEE conference on computer vision and pattern recognition,  2016, pp. 770--778.

\bibitem[Chen et~al.(2023)Chen, Huang, Liu, Wang, Liu, Zhang, Su, and Zhang]{CHEN2023113802}
Chen, P.; Huang, H.; Liu, J.; Wang, J.; Liu, C.; Zhang, N.; Su, M.; Zhang, D.
\newblock {Leveraging Chinese GaoFen-7 imagery for high-resolution building height estimation in multiple cities}.
\newblock {\em Remote Sensing of Environment} {\bf 2023}, {\em 298},~113802.
\newblock {\url{https://doi.org/https://doi.org/10.1016/j.rse.2023.113802}}.

\end{thebibliography}
\end{document}